\documentclass{article}

    \usepackage[preprint]{neurips_2023}

\usepackage[utf8]{inputenc} 
\usepackage[T1]{fontenc}    
\usepackage{url}            
\usepackage{booktabs}       
\usepackage{amsfonts}       
\usepackage{nicefrac}       
\usepackage{microtype}      
\usepackage{xcolor}         
\usepackage{mathrsfs}
\definecolor{linkColor}{rgb}{0.,0.44,0.74}
\definecolor{cgcolor}{rgb}{0,0.6,0.33}
\usepackage[colorlinks,citecolor=linkColor]{hyperref} 
\hypersetup{
hidelinks,
colorlinks=true,
linkcolor=red,
citecolor=cgcolor
}

\usepackage[font=small,labelfont=bf]{caption}
\usepackage{times}
\usepackage{epsfig}
\usepackage{graphicx}
\usepackage{amsmath}
\usepackage{amssymb}
\usepackage{threeparttablex}
\usepackage{xspace}
\usepackage{multirow}
\usepackage{bbding}
\usepackage{multicol}
\usepackage{pgfplots}
\usepackage{appendix}
\usepackage{color, xcolor}
\usepackage{float}
\usepackage[caption=false,font=scriptsize]{subfig}
\usepackage{listings}
\usepackage{colortbl, booktabs}
\usepackage{enumitem}
\usepackage{tablefootnote}
\usepackage{wrapfig}
\definecolor{light}{gray}{.85}
\setcitestyle{numbers,square}
\usepackage{amsmath}

\title{
$\texttt{\textbf{VideoLLM:}}$ Modeling Video Sequence with Large Language Models

}

\author{
  Guo Chen$^{1,2}$, Yin-Dong Zheng$^{1}$, Jiahao Wang$^{1}$, Jilan Xu$^{2,3}$, Yifei Huang$^{2}$, Junting Pan$^{4}$ \\ 
  \textbf{Yi Wang$^{2}$, Yali Wang$^{2}$, Yu Qiao$^{2}$,
  Tong Lu$^{1}$, Limin Wang$^{1,2}$} \\
  $^1$ Nanjing University, 
  $^2$ OpenGVLab, Shanghai AI Laboratory, 
  $^3$ Fudan University\\
  $^4$ The Chinese University of Hong Kong
  \\
  \\
  \textbf{\url{https://github.com/cg1177/VideoLLM}}
}

\begin{document}

\maketitle

\begin{abstract}
With the exponential growth of video data, there is an urgent need for automated technology to analyze and comprehend video content. However, existing video understanding models are often task-specific and lack a comprehensive capability of handling diverse tasks.
The success of large language models (LLMs) like GPT has demonstrated their impressive abilities in sequence causal reasoning. Building upon this insight, we propose a novel framework called $\texttt{\textbf{VideoLLM}}$ that leverages the sequence reasoning capabilities of pre-trained LLMs from natural language processing (NLP) for video sequence understanding.
$\texttt{\textbf{VideoLLM}}$ incorporates a carefully designed Modality Encoder and Semantic Translator, which convert inputs from various modalities into a unified token sequence. This token sequence is then fed into a decoder-only LLM. Subsequently, with the aid of a simple task head, our $\texttt{\textbf{VideoLLM}}$ yields an effective unified framework for different kinds of video understanding tasks.
To evaluate the efficacy of $\texttt{\textbf{VideoLLM}}$, we conduct extensive experiments using multiple LLMs and fine-tuning methods. We evaluate our $\texttt{\textbf{VideoLLM}}$ on eight tasks sourced from four different datasets. The experimental results demonstrate that the understanding and reasoning capabilities of LLMs can be effectively transferred to video understanding tasks.
\end{abstract}

\section{Introduction}
\label{sec:intro}
The advent of phenomenon-level language applications, such as ChatGPT~\cite{openai2022chatgpt}, has showcased LLMs'~\cite{gpt,gpt2,gpt3,gpt4,t5,opt,llama,wang2023visionllm} remarkable zero-shot capability in effectively addressing multiple NLP or vision-centric tasks. The remarkable sequence modeling and reasoning capabilities that these large language models exhibited can be traced back to their acquisition through rigorous pre-training with substantial parameters on large-scale corpora. Despite the amazing achievements in processing language sequences, understanding video sequences that record the real world's objective laws and can be regarded as long image sequences is far from the level of present LLM.

Video sequence understanding involves various real-world applications, such as surveillance systems \cite{surveillance}, autonomous vehicles \cite{soran2015auto}, robotics \cite{robot}, and wearable devices \cite{wearable-device}. 
Simply put, it involves AI systems in the real-time processing of visual information streams, reasoning them in the context of long-term time series, and then providing responses. The vanilla paradigm for video sequence understanding tasks relies on task-specific designs~\cite{lstr,testra,DCAN,asformer,TadTR,zhang2020span,moment_detr_qvhighlights} to encode or decode video sequences, thereby achieving a promising performance but brings additional tailored cost. Compared with natural language, there is no scalable video sequence model that can be seamlessly adapted to different video sequence tasks.
This is primarily attributed to the challenges associated with large-scale video self-supervision, which arise from the expensive nature of temporal-intensive visual annotation, as well as the time-consuming process of acquiring and processing extensive video data. As a result, there is a pressing demand for an efficient method that can offer fundamental modeling capabilities for tasks involving video sequence understanding.

\begin{figure}[t]
  \centering
\includegraphics[width=1.0\textwidth]{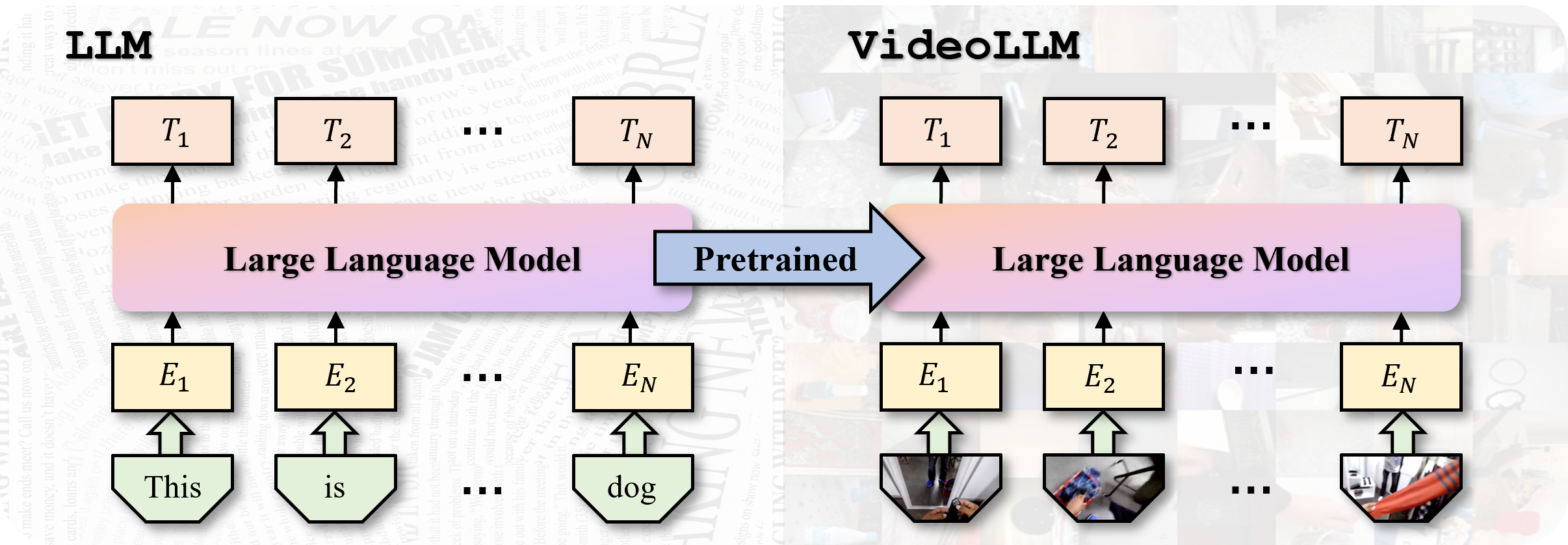}
  \caption{\textbf{Overview of our motivation and method.} (a) LLM taking words as input is pretrained on large-scale nature language composed of word sequences. (b) $\texttt{\textbf{VideoLLM}}$ encodes video stream to token sequences and applies large-scale pre-trained LLMs to video sequence reasoning tasks.}
  \label{fig:overview}
  \vspace{-20pt}
\end{figure}

In this work, we present a novel paradigm called $\texttt{\textbf{VideoLLM}}$, as shown in Figure~\ref{fig:overview}, which aligns video and language sequences and harnesses LLMs' reasoning and understanding capabilities. This paradigm enables videos to engage in reasoning about real-world events through the medium of language. Specifically,
it is composed of three core components: (1) a temporal-wise unitization method to encode unit-wise data stream, (2) an appended semantic translator to transfer visual semantics to language semantics, and (3) a decoder-only LLM as a generalist video sequence reasoner for various video sequence understanding tasks. The design allows sequence tasks with different modalities (\emph{e.g.} visual and text) to be seamlessly integrated, as we verified in the experiments visual-only tasks such as temporal action detection and action anticipation, etc., and visual-language tasks such as temporal grounding and highlight detection, etc. The unit-wise encoding and decoder-only reasoning enable the system to run with minimal delay, greatly meeting real-time or interactive systems' experience requirements.

In contrast to the long-term temporal post-fusion approach proposed in \cite{vivit}, our method emphasizes learning short-term visual token representations for effectively integrating frozen LLMs. This adaptation is conducted within a well-pretrained LLM with robust sequence processing and causal reasoning abilities. Consequently, long-term video modeling can be disregarded, effectively simplifying the complexity of the system design. Compared to recent API-based or ensemble-based visual understanding applications~\cite{chen2022visualgpt,mmreact,shen2023hugginggpt,liu2023internchat,li2023videochat}, we offer an end-to-end system-level approach for video understanding by bridging visual models and LLMs, enhancing the overall efficiency of the long-term video sequence understanding pipeline. Moreover, our method achieves maximal decoupling between short-term and long-term visual modeling, enabling the flexible adoption of heterogeneous short-term visual encoding techniques while rapidly incorporating state-of-the-art LLMs.

Our contributions can be succinctly summarized as follows:

(1) We present $\texttt{\textbf{VideoLLM}}$, a novel framework that harnesses the sequence reasoning capabilities of pre-trained LLMs to tackle video sequence understanding tasks through the medium of language. By aligning videos with language, $\texttt{\textbf{VideoLLM}}$ enables simultaneous reasoning about language logic and the evolution of real-world states through unified modeling.

(2) We reexamine the characteristics and challenges associated with various video sequence understanding tasks and develop a novel, plug-and-play adaptation scheme to adapt off-the-shelf visual encoders and advanced LLMs effectively. This scheme is built upon a unified adaptation principle, eliminating the need for task-specific customization.

(3) We conduct extensive experiments across four datasets, encompassing eight video sequence understanding tasks. These tasks encompass diverse settings, including data accessibility (causal or non-causal), perceptual objectives (memory or anticipation), prediction granularity (segment-level or frame-level), and modalities (vision-only or vision-language). The experiments employ a range of LLMs, such as GPT-2, T5, and OPT.
Comparative analyses against task-specific tailored models demonstrate that our $\texttt{\textbf{VideoLLM}}$ achieves state-of-the-art or comparable performance on these tasks, employing comparable or fewer trainable parameters. These results effectively establish LLM as an effective video reasoner, while validating the efficacy of our proposed $\texttt{\textbf{VideoLLM}}$ framework for multiple video sequence understanding tasks.

\section{Related Work}
\label{sec:rw}

\subsection{Video Sequence Understanding} 
Video Sequence Understanding tasks can be categorized into two types based on the granularity of predictions: timestamp-level tasks and segment-level tasks. Timestamp-level tasks aim to predict closed-set properties at each time step or filter suitable time steps based on textual conditions. For example, \cite{oad2016,oadtr,lstr,ms-tcn,asformer} implement online action detection or action segmentation tasks to predict the category of each time step in a video stream. Similarly, \cite{testra,girdhar2021avt,rulstm,tempagg} implement action anticipation tasks to predict the action category that occurs after a certain time gap. Additionally, methods such as \cite{moment_detr_qvhighlights,liu2022umt} achieve text-based highlight detection.
Segment-level tasks involve predicting segment boundaries in a video sequence based on closed-set categories or open text. Related tasks include moment query \cite{bsn,g-tad,zhao2021vsgn,yang2023basictad,actionformer} and natural language query \cite{zhang2020span,ramakrishnan2023naq,xu2023boundary}. The model proposed in this paper is tested on multiple video sequence understanding tasks to verify the language models' capability to reason about videos from different perspectives.

\subsection{Vision Models}
Vision Models, including image and video models, have recently been developed rapidly, mainly focusing on representing short-term vision information. Vision models are divided into convolution, transformer, and hybrid networks. Convolution models learn spatial~\cite{bninception,resnet,resnext,convnext,yang2022focalnet,wang2022internimage} or space-time~\cite{tsn,i3d,slowfast,r2+1d,csn,tdn,tam} visual representations by aggregating neighborhood information using 2D or 3D convolution operators. With the great success of the transformer~\cite{transformer} in the NLP field, the visual transformer has also been continuously developed. The visual transformer models space~\cite{vit,swin,wang2021pvtv1,deit,beit,eva} or space-time~\cite{mvit,videomae_facebook,timesformer,vivit,videomae,wang2023videomaev2} through an attention mechanism. Due to the data-hungry problem caused by the lack of inductive bias in the transformer network, a hybrid network~\cite{pvtv2,uniformer,ali2021xcit,li2022uniformerv2,wang2022internvideo} combining attention mechanism and convolution operator is proposed to improve performance.

\subsection{Large Language Models}
Large Language Models have emerged in recent years in natural language processing. These models usually contain billions to hundreds of billions of parameters and are trained on large text corpora \cite{gpt,gpt2,flan,t5,chinchilla, palm,llama}. The core architecture of the model is based on the Transformer \cite{transformer} while the objective functions range from masked language modeling \cite{bert,roberta,spanbert}, generative language modeling \cite{gpt,gpt2,gpt3} and permuted language modeling \cite{xlnet}. Among these works, the generative-based language models showed promising results \cite{gpt2,gpt3} on a wide range of natural language understanding benchmarks. Beginning with the representative work GPT-3 \cite{gpt3}, a series of works \cite{Megatron-TuringNLG,gopher,chinchilla,opt,palm,llama} scaled up the model and pre-training data and demonstrated strong few-shot and zero-shot performance. Despite the promising results on natural language tasks, the capability of the models are still less explored in multimodal domain. In this paper, we attempt to discover the long-range modeling capacity of LLMs in improving video understanding.

\subsection{Multimodal Models}
Multimodal Models aim to learn joint vision and language representation for multimodal downstream tasks. The dominant works are VLP models trained end-to-end on large-scale image/video-text pairs \cite{clip,align,albef,clipbert,allinone,frozenintime,lin2022egovlp}. To relieve the high computation resources, modulated vision-language models adopted frozen unimodal or multimodal pre-trained encoders with learnable modules \cite{blip,blip2,flamingo}. These models leveraged strong representation ability of large language models for alignment or generation tasks. 
BLIP-2 \cite{blip2} trained a lightweight Transformer to compress the visual tokens and built a bridge between vision output and language input. Flamingo \cite{flamingo} injected visual features into LLM by adding intermediate cross-attention Transformer layers.

\section{Preliminary}

\subsection{Large Language Model}

\begin{wraptable}{r}{0.38\textwidth}
\vspace{-10pt}
\small
 \begin{tabular}{lrr}
    Model & \#Param & \#Tokens\\
    \midrule[1.5pt]
        GPT-2~\cite{gpt2} & 1.5B & 10B\\
        GPT-3~\cite{gpt3} & 175B & 499B\\
        T5~\cite{t5} & 11B & 156B \\
        OPT~\cite{opt} & 175B & 180B \\
        PaLM~\cite{palm} & 540B & 780B \\
        LaMDA~\cite{thoppilan2022lamda} & 137B & 1.56T \\
        LLaMA~\cite{llama} & 65B & 1.4T \\
    \end{tabular}
        \caption{Parameter and training scale of LLMs.}
    \label{tab:llm-paramter-tokens}
    \vspace{-10pt}
\end{wraptable}

The current Language Model can be mainly sorted into encoder-decoder and decoder-only structures.
The encoder-decoder uses bidirectional Masked Language Modeling to restore corrupted tokens in a document for textual representation learning, such as BERT~\cite{bert} and T5~\cite{t5}.
Alternatively, the decoder-only (GPT family~\cite{gpt}, OPT~\cite{opt}) uses unidirectional Language Modeling to directly maximize the likelihood of the sequence under the forward autoregressive factorization.
These two training mechanisms grant the language model powerful language sequence modeling and reasoning capabilities.
Model parameters and data size of Language models are continuous growth.
Table~\ref{tab:llm-paramter-tokens} lists the model parameter amount and pre-training token size.
These models usually adopt different network structures, training strategies, and corpora.
We will explore various LLMs' performance, advantages, and drawbacks as video sequence reasoners.

\subsection{Tasks}
$\texttt{\textbf{VideoLLM}}$ is verified on 8 video understanding tasks across 4 datasets in Table~\ref{tab:dataset-and-tasks}. \textbf{Online Action Detection}, \textbf{Action Segmentation}, and \textbf{Temporal Action Detection} focus on detecting and recognizing actions and their temporal boundaries. \textbf{Online Captioning} generates textual descriptions of video content, while \textbf{Highlight Detection} identifies exciting parts and generates summaries. \textbf{Action Anticipation} and \textbf{Long-term Anticipation} predict future actions and content in advance, respectively. \textbf{Moment Query} quickly retrieves specific segments or events in a video. \textbf{Nature Language Query} localize a temporal segment through a textual question.

\begin{table}[h]
\centering
\small

\begin{tabular}{lll}
Task & Datasets & Metric \\
\midrule[1.5pt]
Online Action Detection & EK100~\cite{epic-Kitchens}  & Recall Top-5 \\
Action Segmentation & Breakfast~\cite{breakfast}  & F1; Edit distance \\
Online Captioning & Ego4D-Narration~\cite{grauman2022ego4d}  & METEOR; ROUGE-L \\
Action Anticipation & EK100~
\cite{epic-Kitchens}& Recall Top-5 \\
Long-term Anticipation & Ego4D-LTA~\cite{grauman2022ego4d}  &  Edit distance \\
Moment Query & Ego4D-MQ~\cite{grauman2022ego4d}  & mAP@IoU \\
Nature Language Query  & Ego4D-NLQ~\cite{grauman2022ego4d}  & Rank@1, Rank@5 \\
Highlight Detection & QVHighlights~\cite{moment_detr_qvhighlights}  & mAP \\
\end{tabular}
\caption{Statistics of datasets in our experiments.}
\label{tab:dataset-and-tasks}
\vspace{-20pt}
\end{table}

\section{\texorpdfstring{$\texttt{\textbf{VideoLLM}}$}{VideoLLM}}
\label{sec:method}

$\texttt{\textbf{VideoLLM}}$ is a novel online video reasoning system that aims to apply large-scale pre-trained Large Language Models to video sequence understanding tasks through parameter-efficient transfer learning. 
It directly borrows the sequence modeling ability of LLM to video sequence reasoning, allowing vision to flow in a natural time sequence in the form of language.

\begin{figure*}[t]  \centering
  \includegraphics[width=1\textwidth]{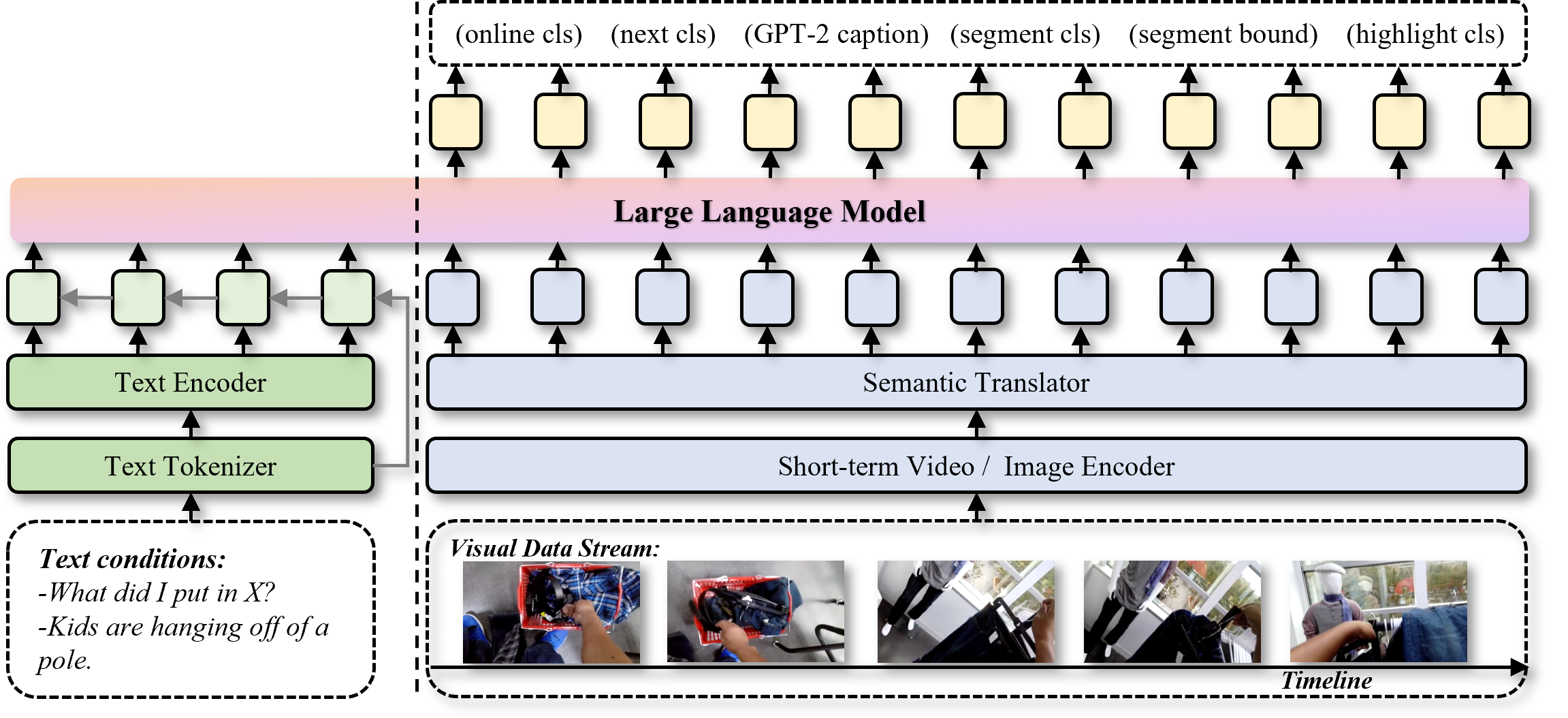}
  \caption{
$\texttt{\textbf{VideoLLM}}$ leverages LLM as its core to handle video and text sequences seamlessly.
In detail, all input video frames are converted into a visual encoding sequence using a short-term visual encoder. On the other hand, the text condition is transformed into a textual sequence using a text encoder or a text tokenizer.
Subsequently, the semantic translator aligns the visual and text encoding, thus feeding the two sequences to LLM for seamless sequence reasoning.
Finally, the output generated by LLM can be applied to various video understanding tasks.}
  \label{fig:framework}
  \vspace{-10pt}
\end{figure*}

This section will overview the $\texttt{\textbf{VideoLLM}}$ architecture, as shown in Figure~\ref{fig:framework}. Specifically, $\texttt{\textbf{VideoLLM}}$ comprises several components: Modality Encoder, Semantic Translator, decoder-only Reasoner, and simple task heads. In this framework, each short video clip is tokenized using corresponding audio and video encoders and then sequentially processed by the LLM. It is important to note that our unified LLM naturally integrates textual conditions into the framework. Furthermore, our framework allows for the easy integration of various human prompts, commands, human-computer interaction techniques, and parameter-efficient fine-tuning techniques to improve model performance and efficiency.

\subsection{Modality Encoder}
We adopt a temporal-wise unitization method to process unit-wise visual (or audio and other modality) information for utilizing LLMs to understand video streams comprehensively.
We naturally consider integrating natural language modeling with LLMs for unified processing to achieve multimodal understanding.

\textbf{Vision.} 
To encode a video sequence of $F$ frames $x \in \mathbb{R}^{F\times H\times W \times C}$ where $H$, $W$, and $C$ are the height, width, and the number of channels of each frame, we use a short-term visual encoder $f_v$, which can be a well-established image encoder or a short-term video encoder.
Given $F_s$ presenting the number of frames in a short-term clip, all frames are divided into $N_v=\frac{F}{F_s}$ space-time visual unit, and each unit is encoded by $f_v$ independently.
Hence, $f_v$ outputs a sequence of space-time visual units $x_v = f_v(x) \in \mathbb{R}^{N_v \times \frac{F_s}{s_t} \times \frac{H}{s_h}\times \frac{W}{s_w} \times d}=\{x_v^1,x_v^2,...,x_v^{N_v}\}$, where $d$ is the representation dimension and $s_t$, $s_h$ and $s_w$ are the strides of space-time dimensions within $f_v$.

\textbf{Text.} We support two encoding approaches when presented with a textual input $y$ containing narration or a question.
The first approach involves tokenizing $y$ into $y_t \in \mathbb{R}^{N_t \times d}$, where $d$ represents the output dimension of the tokenizer. The other is to process further $y_t$ using language encoders $f_t$, such as BERT~\cite{bert}, T5~\cite{t5}, or CLIP~\cite{clip}, to extract textual features denoted as $y_e$. Subsequently, either $y_t$ or $y_e$ can be employed as input for the video sequence reasoner to implement the control based on text condition.

\subsection{Semantic Translator} 

The language model is essentially a blind who can receive language input and learn various knowledge, but it has no vision and cannot directly perceive the visual world. 
Therefore, we need to translate the visual semantics into language representations that the language model can interpret.

Similar to Perceiver~\cite{jaegle2021perceiver}, Flamingo~\cite{flamingo}, and BLIP-2~\cite{blip2}, we adopt an appended sub-network to transfer the semantic space. In this paper, for efficiency, we adopt a simpler design that freezes the visual encoder and transfers the final visual feature into the language world. In detail, given $x_v \in \mathbb{R}^{N_v \times \frac{F_s}{s_t} \times \frac{H}{s_h}\times \frac{W}{s_w} \times d_v}$, we first pool each visual unit of $x_v$ to the temporal token. Hence, we obtain a video sequence representation $x_t \in  \mathbb{R}^{N_v \times d_v}$. We use one linear projector $\phi$ to learn translation from the visual to language semantics to attain translated semantics $s_v=\phi(x_t) \in \mathbb{R}^{N_v \times d}$, where $d$ is the hidden dimension of the used LLM.

\subsection{Decoder-only Reasoner} 
\begin{figure*}[t]  \centering
  \includegraphics[width=1\textwidth]{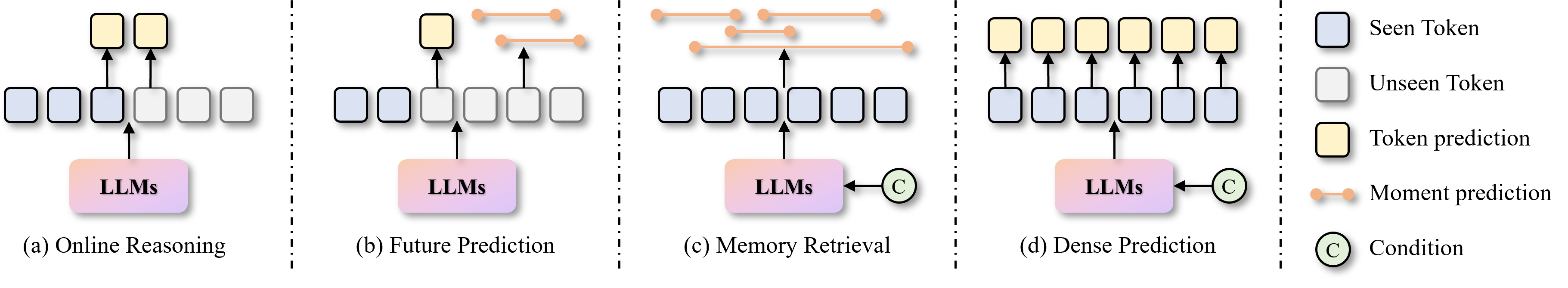}
  \caption{\textbf{Schematic diagram of LLMs adapting to 4 types of tasks.} (a) ``Seen Tokens'' denote the data units accepted by the AI system and encoded by the encoder. Predicting the attributes of the latest seen token or near-term unseen token can be seen as an online reasoning task.
  (b) ``Unssen tokens'' are data units that have not yet arrived, and predicting their attributes or when they appear in the future usually belongs to future prediction tasks. (c) Given a text condition or a closed category set, retrieving ``moments'' from a past sequence of seen tokens, also known as memory, is a memory retrieval task. (d) A similar task for memory called dense prediction predicts attributes of each seen token or highlights tokens that match the condition.
  }
  \label{fig:adapt-4task}
  \vspace{-10pt}
\end{figure*}

As detailed in Table~\ref{tab:dataset-and-tasks}, our objective is to enable our $\texttt{\textbf{VideoLLM}}$ to accommodate a broad range of video sequence understanding tasks. However, the disparate constraints inherent to these tasks, including their respective inputs and outputs, are a potential obstacle to achieving this goal.
To better understand the multifaceted nature of these tasks, we have classified them into four categories, which may exhibit some overlap, as illustrated in Figure~\ref{fig:adapt-4task}.
This section will discuss efficiently adapting LLMs to address different video understanding tasks.

We employ LLM with a decoder-only structure, denoted as $\mathcal{M}$, as the key component of our video sequence reasoner, informed by three critical considerations.
First, compelling evidence indicates that decoder-only LLMs are particularly adept at handling causal reasoning tasks for language sequences.
Second, the most advanced and high-performing large language models in the current landscape are predominantly decoder-only and are subject to continuous optimization by the research community.
Third, a real-world video processor should ideally be designed around a unidirectional visual data flow to maximize performance. This design philosophy aligns seamlessly with the underlying structure of decoder-only language models.
Subsequently, we provide a succinct overview of our adaptation method.

\textbf{Online Reasoning.} Online Reasoning primarily focuses on real-time prediction of the category or caption for the most recently attended data unit, which in this paper refers to a new short-term video clip.
Given a playing video stream and working memory $m=\{s_v^{-t+1},s_v^{-t+2},...,s_v^{i},...,s_v^{0} \}$, where $t$ is the number of seen tokens in memory and $s_v^0$ is the latest translated token. In the training phase, $m$ will be fed into $\mathcal{M}$ to construct a causal sequence $c=\{c^{-t+1},c^{-t+2},...,c^{i},...,c^{0} \}$ for parallel training. We use two linear layers to predict the category of each token $s_v^{i}$ and its next token $s_v^{i+1}$. Thanks to the causal structure of decoder-only LLM, we do not need to calculate the context of the entire sequence when accepting a novel token in the inference phase, compared with a bidirectional encoder. We only make $s_v^0$ cross-attend to the historical context to calculate new states $c^{0}$. Additionally, we use each $c^i$ as the hidden states for online captioning and input into an extra generative language model $\mathcal{M}_g$ (e.g., GPT-2~\cite{gpt2}) for autoregressive text generation.

\textbf{Future Prediction.} Given a sequence of seen tokens $m=\{s_v^{-t+1},s_v^{-t+2},...,s_v^{i},...,s_v^{0} \}$ as the working memory, model need predict the next $N_{f}$ tokens or  events. In this case, we still utilize the causal structure, supervising each seen token to learn future representations. For predicting different $N_{f}$ future states, we use $N_{f}$ normalization layers to separate $N_{f}$ anticipation presentations $a=\{a^{1},a^{2},...,a^{i},...,a^{N_f} \}$. 

\textbf{Memory Retrieval.} Memory Retrieval often is an offline task to detect event segments in a closed category set or by a text condition. In our online system, however, the task can evaluate the model's understanding of segment-level transitions and evolutions in the video. Given a sequence of seen tokens $m=\{s_v^{-t+1},s_v^{-t+2},...,s_v^{i},...,s_v^{0} \}$ as the working memory, to get the context of the whole video, we use the last token $s_v^{0}$ to predict segments in the memory. Another alternative is to concatenate a learnable token $s_v^q$ or $\texttt{<EOT>}$ at the end of the $m$ to learn the memory summary. To predict at most $N_m$ possible segments with category-closed in memory, similar to future prediction, we use $N_m$ normalization layers to separate $N_m$ segment-level memory presentations $m_s=\{m_s^{1},m_s^{2},...,m_s^{i},...,m_s^{N_m} \}$. Then we adopt two linear layers to predict the category and boundary of 
each segment. The segments are matched with ground truth through Hungarian matching algorithm~\cite{detr} for supervision. For memory retrieval based on text condition, we concatenate text presentation $y_t$ or $y_e$ at the end of $m$ and feed them into $\mathcal{M}$ together. Hence, $\mathcal{M}$ can generate the causal sequence conditioned on text for retrieving matched moments.

\textbf{Dense Prediction.}
Dense Prediction can be likened to an offline reasoning task where the goal is to predict the category of each token or identify highlight tokens based on textual conditions. In this work, we treat dense prediction as an online task, which serves as a simplified implementation of online action segmentation or highlight detection. 
Our system uses decoder-only LLM as the default video reasoner and handles online prediction and text conditions like the aforementioned tasks.
However, it is worth exploring whether a bidirectional reasoner can provide performance improvements for memory-related tasks. Therefore, we also consider a bidirectional encoder as a potential candidate for our video reasoner, which we evaluate in subsequent experiments.

In summary, our experimental objective is to assess the \emph{intrinsic capability} of $\mathcal{M}$ in understanding video sequences. To accomplish this, we propose three fundamental adaptation principles, which have been adhered to by the aforementioned methods. Firstly, we exclusively supervise tasks by relying on the final output of $\mathcal{M}$, instead of employing multi-stage supervision as demonstrated in the works of \cite{TadTR} and \cite{asformer}. Secondly, we refrain from incorporating prior operators, such as convolution layers, into $\mathcal{M}$. Lastly, we employ linear layers for each task to transform the hidden states generated by $\mathcal{M}$ into task results, thereby eschewing the utilization of intricate task-specific heads.

\subsection{Model Training}
The training process of $\texttt{\textbf{VideoLLM}}$ involves three fine-tuning methods for training the model.

\textbf{Basic Tuning.} When working with a frozen language model, the optimization of $\texttt{\textbf{VideoLLM}}$ primarily focuses on fine-tuning the semantic translation and output layers. In this scenario, the model's performance completely relies on the capabilities of the LLM after semantic translation.

\textbf{Partial Tuning.} The partial tuning method involves optimizing specific parts of the LLM in addition to the basic tuning. We adopt three settings for partial tuning: optimizing all bias parameters, optimizing the first block, and optimizing the last block.

\textbf{PEFT Tuning.}  
The widely popular and effective parameter-efficient fine-tuning (PEFT) techniques in NLP, such as LoRA~\cite{lora}, Prompt Tuning~\cite{prompt-tuning}, and Prefix Tuning~\cite{prefix-tuning}, have also been applied to optimize $\texttt{\textbf{VideoLLM}}$.

\section{Experiments}
\label{sec:experiments}

\subsection{Experimental Setup}

\textbf{Dataset and Tasks.}
In order to thoroughly assess the capabilities of LLMs in video understanding, we performed experiments on four datasets, covering a total of eight tasks. The details of these tasks and datasets are presented in Table~\ref{tab:dataset-and-tasks}. The tasks were categorized into four types, as illustrated in Figure~\ref{fig:adapt-4task}: Online Reasoning, Future Prediction, Memory Retrieval, and Dense Prediction. This diverse set of tasks allows for comprehensive evaluations from various perspectives, including data accessibility (causal or non-causal), perceptual objectives (memory or anticipation), and prediction granularity (segment-level or frame-level), modalities (vision-only or vision-language).

\textbf{Evaluation and Metrics.} Our model evaluation is conducted in accordance with previous studies~\cite{oad2016,testra,ms-tcn,anet,grauman2022ego4d,moment_detr_qvhighlights,chen2022internvideo_ego4d}.
Specifically, we measure the accuracy of online action detection and action anticipation tasks using class-mean recall@5(\%) following the established standard protocol~\cite{epic-Kitchens}.
To assess the performance of our model in the action segmentation task, we report the framewise accuracy (Acc), segmental edit distance (ED), and the segmental F1 score at overlapping thresholds of 25\% denoted as F1@25.
For the Long-term anticipation task, we submit our results to the EvalAI platform to evaluate the \emph{test} set. 
Consistent with the approach employed in~\cite{grauman2022ego4d}, we evaluate the mean Average Precision (mAP) under multiple temporal Intersection over Union (tIoU) thresholds, specifically $\{0.1; 0.2; 0.3; 0.4; 0.5\}$, for the Moment Query task. In addition, we report the recall@k, where k = 1, and the IoU=m metric, where m = $\{0.3, 0.5\}$, for the Nature Language Query task.

\textbf{Implementation Details.}
To ensure fairness and facilitate meaningful comparisons within the research community, we employ various visual encoders~\cite{i3d,clip,slowfast,2stream_tsn,videomae,lavila,wang2022internvideo} that have been pretrained on different datasets~\cite{deng2009imagenet,kinetics,clip,grauman2022ego4d,epic-Kitchens} to extract visual features. This approach helps establish alignment with existing community settings and ensures equitable evaluations.
Note that, the same modality encoder could share semantic translator. In this work, using different encoders and semantic translators for aligning community settings is a special case.
In particular, we adopt the fundamental settings proposed in~\cite{lstr,testra} for the Online Action Detection and Action Anticipation tasks. We leverage the settings introduced in~\cite{asformer} for the Action Segmentation task. The Online Captioning task follows the settings outlined in~\cite{lavila}. Similarly, we adhere to the settings specified in~\cite{grauman2022ego4d} for the Long-term Anticipation, Moment Query and Nature Language Query task. The Highlight Detection task builds upon the settings presented in~\cite{moment_detr_qvhighlights}.


\subsection{Main Results and Analysis}

\begin{figure*}[t]  
\centering
  \includegraphics[width=\textwidth]{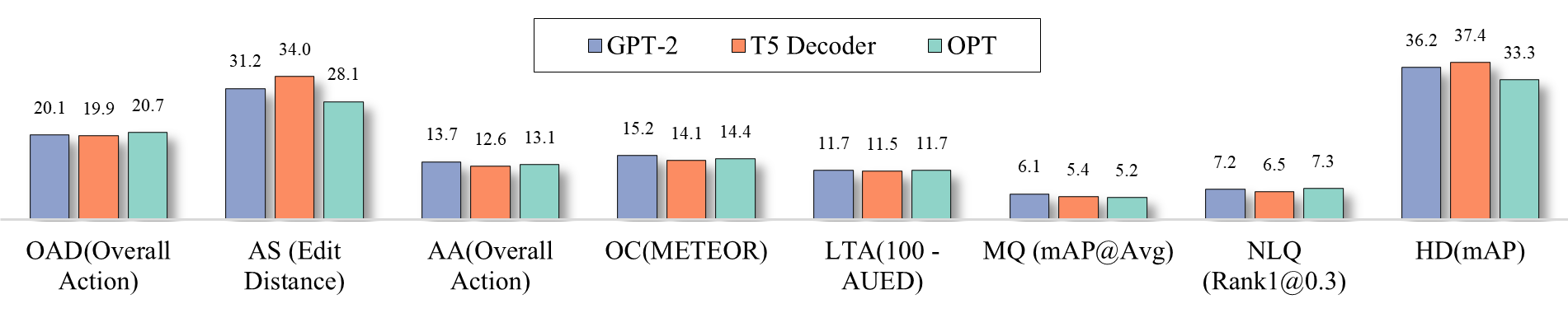}
  \caption{We conducted a performance comparison of different base-level Language Models with basic tuning across various tasks. We compared the performance of GPT-2~\cite{gpt2}, T5 Decoder~\cite{t5}, and OPT~\cite{opt}. For each task, we select representative metrics to facilitate the comparison. For LTA task, we report the results on \emph{val} set.}
  \label{fig:llm-better}
  \vspace{-20pt}
\end{figure*}

\textbf{Which language model performs better?}
Figure~\ref{fig:llm-better} presents the comparison results between three base-level LMs, GPT-2~\cite{gpt2}, T5 Decoder~\cite{t5}, and OPT~\cite{opt}. 
The results are obtained through the basic tuning method. 
We select representative metrics for each task for intuitive comparison. 
From the results, we can see that different language models have different performances on different video sequence understanding tasks. Both GPT-2 and OPT are better than T5 decoder in future prediction tasks (see AA and LTA in the figure). On the contrary, OPT is significantly better than GPT-2 and T5 decoder in OAD task. For Moment Retrieval tasks, we find that GPT-2 can still gain dominance (see MQ and NLQ in the figure). It is worth noting that T5 Decoder has a great advantage over GPT-2 and OPT in dense prediction tasks (see AS and HD in the figure). For online captioning, GPT-2 attains the best performance, compared with T5 Decoder and OPT. We suppose that using GPT-2 as video sequence reasoner $\mathcal{M}$ better aligns the text generator $\mathcal{M}_g$ (also GPT-2) we used from~\cite{lavila}.
In general, the structure and training strategy of the language model will result in different processing capabilities for video sequences and exhibit different adept abilities. In fact, when we calculated their average scores based on the results, we found that GPT-2 and T5 decoder were basically on par, and OPT was slightly worse than GPT-2 and T5 decoder.

\begin{wraptable}{r}{0.5\textwidth}
\vspace{-10pt}
\scriptsize
\setlength{\tabcolsep}{1.0mm}
    \begin{tabular}{lll}
        Model & Action Top-5 Recall   & +Trainable Param (M) \\
        \midrule[1.5pt]
        
        Basic & 20.1  &  0  \\
        LoRA ($r$=1/2/4/8) & 19.5/19.7/19.8/19.6 & 0.04/0.07/0.15/0.30 \\
        Prompt ($r$=1/2/4/8) & 20.3/20.6/20.7/20.8&  0.00/0.00/0.00/0.00 \\
        Prefix ($r$=1/2/4/8) & 20.8/20.6/\textbf{21.4}/21.1& 0.02/0.04/0.07/0.15 \\
        Partial (bias/F/L/FL) & 20.5/20.6/20.5/20.8& 0.1/7.09/7.09/14.18 \\
    \end{tabular}
\caption{Impact of different tuning methods using GPT-2 on OAD task. $r$ denote the hyperparameter of the three PEFT tuning methods. ``F'' and ``L'' represent the first block and the last block in LMs. }
\label{tab:tuning-method}
\vspace{-10pt}
\end{wraptable}

\textbf{Which Tuning method performs better?} To evaluate the influence of various tuning methods on performance, we opt OAD as the experimental object. It is a causal dense prediction task, providing a more realistic representation of performance alterations.
Table ~\ref{tab:tuning-method} presents the Action Top-5 Recall achieved through the utilization of various tuning methods, along with the corresponding increase in trainable parameters compared to the basic tuning approach.
We employ $r$ as a uniform representation of the hyperparameter for the three PEFT tuning methods, and carry out experiments using $r=1/2/4/8$. As depicted in the table, employing LoRA with different $r$ results in a decline in performance. Conversely, the other tuning methods exhibit performance improvements of at least 0.2 points in the Action Top-5 Recall metric.
Although fine-tuning the first or last block can yield performance gains, it also entails a significantly larger number of trainable parameters compared to the other methods.
Remarkably, when employing prefix tuning with $r=4$, the model achieves the best outcome, attaining an Action Top-5 Recall of 21.4, surpassing the basic tuning method by 1.3 points.

\textbf{Comparison to the state-of-the-art methods.}
Table~\ref{tab:sota} presents the evaluation results for seven video sequence understanding tasks. It is important to note that the OC task is not included in this analysis due to the lack of comparable sequence-level methods. To thoroughly assess the effectiveness of $\texttt{\textbf{VideoLLM}}$, we conduct a comparative analysis with other cutting-edge methods that are specifically tailored to individual tasks. The reported results for $\texttt{\textbf{VideoLLM}}$ represent the most favorable performance achieved from numerous combinations.
To evaluate the OAD task, we reproduce the existing state-of-the-art methods~\cite{lstr,testra} and adopt the same evaluation metrics~\cite{epic-Kitchens} as the AA task. Notably, we ensure a fair comparison by excluding the data augmentation techniques employed by Testra~\cite{testra}.
Our model demonstrates higher or comparable performance in both the OAD and AA tasks. Particularly, our approach achieves a higher Unseen Action Top-5 Recall, highlighting the ability of utilizing LLMs to ensure and potentially enhance generalization in unseen scenarios. For the AS task, our model outperforms the state-of-the-art method MS-TCN~\cite{ms-tcn} in terms of F1@25, edit distance, and accuracy. It is worth emphasizing that our adaptation principle solely relies on the sequence modeling capability of the LMs themself, without introducing any local prior operator or multi-stage refinement. This observation emphasizes that a language sequence-trained model can serve as a robust initialization for video sequence modeling. We also apply our adaptation principles to MS-TCN~\cite{ms-tcn} and ASFormer~\cite{asformer}, with the corresponding results presented in the table. In the table, SS-TCN$^{\dagger}$ refers to the deep network with a single-stage supervision mentioned in the MS-TCN paper. These results demonstrate a significant inferiority to our single-stage adaptation.
Furthermore, we compare $\texttt{\textbf{VideoLLM}}$ against state-of-the-art or baseline methods on multiple sub-tasks, namely LTA, MQ, and NLQ, of Ego4D~\cite{grauman2022ego4d}. The evaluation conducted on the LTAv2 \emph{test} set, using the EvalAI platform, shows that our model outperforms the official baseline methods. Moreover, under the constraints of the adaptation principle, our model exhibits a slight performance superiority over VSGN~\cite{zhao2021vsgn}, which employs an anchor-based prior setting for the MQ task.
In the realm of visual-language tasks, our models exhibit substantial superiority over existing state-of-the-art methods~\cite{chen2022internvideo_ego4d,moment_detr_qvhighlights}. This finding underscores the impressive performance of language models once the vision-to-language semantic translation is accomplished. 
Furthermore, in addition to the performance comparisons, we also compare the trainable parameters with these methods. The table reveals that our method necessitates approximately 2M to 15M learnable parameters across multiple tasks, with most of these parameters primarily utilized in semantic translator and task head. This substantiates the parameter efficiency of our proposed framework.
In summary, these results convincingly demonstrate the adaptability of our proposed framework across a diverse range of video sequence understanding tasks, each with its own unique settings.

\begin{table}[t]
\centering
\scriptsize
\setlength{\tabcolsep}{0.85mm}
    \begin{tabular}{lccccccccccccccccccc}
    \multirow{2}{*}{Model} & \multirow{2}{*}{\begin{tabular}[c]{@{}c@{}}Trainable\\ Param\end{tabular}} & \multicolumn{3}{c}{OAD} & \multicolumn{3}{c}{AA} & \multicolumn{3}{c}{AS} & \multicolumn{3}{c}{LTA} & \multicolumn{1}{c}{MQ} & \multicolumn{2}{c}{NLQ}& \multicolumn{1}{c}{HD} \\ 
    \cmidrule(lr){3-5} \cmidrule(lr){6-8} \cmidrule(lr){9-11} \cmidrule(lr){12-14} \cmidrule(lr){15-15} \cmidrule(lr){16-17}  \cmidrule(lr){18-18} 
     &  & O & U & T & O & U & T & F1@25 & ED & Acc & V$\downarrow$ & N$\downarrow$ & A$\downarrow$ & mAP & R1@0.3 & R1@0.5 & mAP \\ 
    \midrule[1.5pt]
    \multicolumn{17}{l}{\emph{non-language-model-based method}} \\
    LSTR$^\ddagger$~\cite{lstr} & 27.19M & 22.6 & 18.7  & 20.7 & - & - & - & - & - & - & - & - & - & - & - & - & - \\
    Testra$^\ddagger$~\cite{testra} & 27.70M & 23.2 & 19.0  & 20.9 & \textbf{15.5} & 12.4 & 11.9 & - & - & - & - & - & - & - & - & - & - \\ 
    ASFormer$^\dagger$~\cite{asformer} & 1.13M & - & -  & - & - & - & - & 27.3 & 16.2 & 31.0 & - & - & - & - & - & - & - \\
    SS-TCN$^\dagger$~\cite{ms-tcn} & 0.80M & - & -  & - & - & - & - & 20.2 & 14.3 & 56.5 & - & - & - & - & - & - & - \\
    MS-TCN~\cite{ms-tcn} & 0.80M & - & -  & - & - & - & - & 52.9 & 61.4 & 65.1 & - & - & - & - & - & - & - \\
    Ego4D~\cite{grauman2022ego4d} & 30.33M & - & -  & - & - & - & - & - & - & - & \textbf{71.7} & 73.6 & 92.5 & - & - & - & - \\
    VSGN~\cite{TadTR} & 3.80M & - & -  & - & - & - & - & - & - & - & - & - & - & 6.03 & - & - & - \\
    InternVideo~\cite{chen2022internvideo_ego4d} & 6.72M & - & -  & - & - & - & - & - & - & - & - & - & - & - & 14.4 & 9.60 & - \\
    Moment DETR~\cite{moment_detr_qvhighlights} & 2.56M & - & -  & - & - & - & - & - & - & - & - & - & - & - & - & - & 36.5 \\
    
    \midrule
    \multicolumn{17}{l}{\emph{language-model-based method}} \\
    \rowcolor{light}
    $\texttt{\textbf{VideoLLM}}$ &  2 - 15M & \textbf{23.4} & \textbf{20.2} & \textbf{21.6} & 15.4 & \textbf{12.6} & \textbf{12.0} & \textbf{55.3} & \textbf{63.4} & \textbf{65.7} & 72.1 & \textbf{72.5} & \textbf{92.1} & \textbf{6.09} & \textbf{15.5} & \textbf{10.1} & \textbf{37.7}\\
    
    \end{tabular}
    \vspace{1em}
\caption{
\textbf{Comparison with the state-of-the-art models} on 7 video sequence understanding tasks.
For OAD and AA tasks, we evaluate \textbf{O}verall, \textbf{U}nseen and \textbf{T}ail Action Top-5 Recall. We follow~\cite{grauman2022ego4d} to evaluate the LTA task with edit distance (ED) of \textbf{V}erb, \textbf{N}oun and \textbf{A}ction on the \emph{test} set, and other tasks are evaluated on \emph{validation} set. We compare performance through the average mAP of tIoU thresholds between 0.1 and 0.5. $^\ddagger$ denote the results we reproduced. $^\dagger$ denotes the results that we align the method with our adaption principle.
}
\label{tab:sota}
\vspace{-30pt}
\end{table}

\begin{wrapfigure}{r}{0.38\textwidth}
\vspace{-10pt}
\begin{tikzpicture}[scale=0.5] 
\begin{axis}[
    xlabel=Parameters (M), 
    ylabel=Recall Top5 Action, 
    tick align=outside, 
    legend style={at={(0.5,-0.2)},anchor=north,legend columns=3} 
    ]
\addplot[line width=2pt,mark=*,blue] plot coordinates { 
    (117,20.1)
    (345,20.2)
    (762,20.7)
    (1542,21.2)
};
\addlegendentry{GPT-2}
\addplot[line width=2pt,mark=triangle,red] plot coordinates {
    (137.9,19.9)
    (435.6,21.0)
    (1643.6,21.4)
};
\addlegendentry{T5 Decoder}
\addplot[line width=2pt,mark=square,cyan] plot coordinates {
    (135,20.3)
    (350,20.5)
    (1300,23.4)
    (2700,22.5)
};
\addlegendentry{OPT}
\end{axis}
\end{tikzpicture}
\caption{Performance of GPT-2, T5 Decoder and OPT with different number of total parameters.}
\label{fig:scaling-up}
\vspace{-10pt}
\end{wrapfigure}
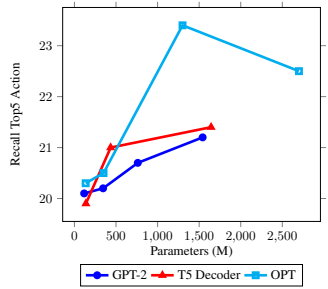

\textbf{Scale of LLM.} We also assess the scalability of utilizing LLMs as video sequence reasoners for our approach, through experimental evaluations conducted on the OAD task.
Figure \ref{fig:scaling-up} displays the Action Top-5 Recall achieved by employing LLMs with varying scales of total parameters. In these experiments, we scale up three decoder-only LLMs, namely GPT-2, T5 Decoder, and OPT, and solely fine-tune two projectors using the basic tuning method. This ensures a comprehensive evaluation of the \emph{intrinsic capabilities} possessed by these LLMs.
As depicted in the figure, when utilizing language models with parameter sizes less than 2B, compelling evidence suggests that larger models yield more substantial improvements in video sequence reasoning. Among the three models, it is worth noting that OPT-1.3B yields the most favorable results, achieving a remarkable 23.4 Action Top-5 Recall. Furthermore, when considering the overall performance improvement trend observed during the scaling-up process, it becomes evident that OPT outperforms T5 Decoder, which, in turn, surpasses GPT-2. However, for larger LLMs, their performance begins to decline. One plausible explanation for this phenomenon is that the dimension-expansion projector causes the model to overfit, as the dimension of the extracted feature sequence is typically less than 2048.
In conclusion, these experiments effectively demonstrate the scalability of our method to LLMs, highlighting their potential for adapting video sequence reasoning tasks.

\begin{wraptable}{r}{0.49\textwidth}
\vspace{-15pt}
\small
 \begin{tabular}{lrccc}
    Model & \#Param & O & U & T\\
    \midrule[1.5pt]
        OPT~\cite{opt} & 6.7B & \textbf{22.1} & 19.9 & \textbf{21.6} \\
        T5 Decoder~\cite{t5} & 6.5B & 19.8 & \textbf{20.2} & 21.1 \\ 
        LLaMA~\cite{llama} & 7B & 21.8 & 20.1 & 21.1 \\
    \end{tabular}
        \caption{Performance of larger and advanced language models, \emph{i.e} OPT-6.7B, T5-Decoder-6.5B and LLaMA-7B on OAD task.}
    \label{tab:advanced-llm}
    \vspace{-10pt}
\end{wraptable}

\textbf{Advanced LLM.}
We further scale up OPT and T5 decoder to 6.7B and utilize the latest 7B LLaMA~\cite{llama} model. The performance of T5 and OPT, as depicted in Table~\ref{tab:advanced-llm}, continues to align with the declining trend observed in Figure~\ref{fig:scaling-up}. Notably, the performance of LLaMA closely approximates that of OPT.

\textbf{Encoder \emph{vs.} Decoder.} We conducted experiments to compare the performance of bidirectional and unidirectional sequence reasoners on three tasks: AS, HD , and NLQ.
For the bidirectional sequence reasoner, we employed the T5~\cite{t5} encoder, while the unidirectional sequence reasoner utilized the T5 decoder. A comprehensive comparison of all task metrics is presented in Table~\ref{tab:encoder-vs-decoder}.

\begin{wraptable}{r}{0.4\textwidth}
\vspace{-15pt}
\centering
\small
\setlength{\tabcolsep}{1.0mm}
    \begin{tabular}{llcc}
        Task & Metric   & Decoder & Encoder \\
        \midrule[1.5pt]
        \multirow{3}{*}{AS} & F1@25 & 25.3 & \textbf{51.1} \\ 
         & ED & 34.0 & \textbf{55.7} \\ 
         & Acc & 44.0 & \textbf{60.7} \\ 
        \midrule
         \multirow{2}{*}{HD} & mAP & 37.4 & \textbf{37.7} \\ 
         & HiT@1 & 61.0 & \textbf{61.6} \\ 
         \midrule
         \multirow{3}{*}{NLQ} & Rank1@0.3 & 6.5 & \textbf{7.4} \\ 
         & Rank1@0.5 & \textbf{3.6} & 3.5 \\ 
         & Rank1@Mean & 5.1 & \textbf{5.5 } \\ 
        
    \end{tabular}
\caption{Impact of encoder and decoder as video sequence reasoner on AS, HD and NLQ tasks. Here encoder and decoder are T5~\cite{t5}.}
\label{tab:encoder-vs-decoder}
\vspace{-10pt}
\end{wraptable}

As evident from the table, the bidirectional reasoner consistently outperformed the unidirectional reasoner in most cases. This discrepancy is particularly prominent in AS tasks, where the bidirectional reasoner exhibited a significantly higher level of performance compared to its unidirectional counterpart. This may be attributed to the importance of bidirectional attention in confirming temporal correlations and pre-post-action relationships within a complete event during action segmentation.
In the case of visual-language tasks, HD and NLQ, the bidirectional reasoner also showcased a slight advantage over the unidirectional reasoner. However, it is worth noting that the Rank1@0.3 obtained by the OPT  on the NLQ task, as depicted in Figure~\ref{fig:llm-better}, is comparable to that achieved by the T5 Encoder (7.3 \emph{vs} 7.4). This suggests that the decoder-only unidirectional reasoner holds the potential to achieve performance on par with the bidirectional reasoner.

\section{Conclusion and Future Work}
In this paper, we propose a novel video understanding framework called $\texttt{\textbf{VideoLLM}}$, which transfers the sequence causal reasoning abilities of large language models (LLMs) from natural language processing to video understanding.
The $\texttt{\textbf{VideoLLM}}$ framework comprises a well-designed Modality Encoder and a Semantic Translator, which convert inputs from different modalities into a unified token sequence.
This sequence is then fed into a decoder-only reasoner realized by the large-scale language pretrained and parameter-frozen LLM, which possesses the ability to decode and output meaningful high-level semantics.
With the help of simple task heads, the output of the LLM corresponds to various specific video understanding tasks.
Extensive experiments were conducted on eight tasks from four different datasets using multiple LLMs and fine-tuning methods to evaluate the effectiveness of $\texttt{\textbf{VideoLLM}}$.
The experimental results demonstrate that LLMs' comprehension and reasoning abilities can be effectively applied to video understanding tasks.
In our future work, we will further explore the potential of LLM. Building upon time series reasoning, we aim to incorporate serialized information about the appearance of video frames, enabling LLM to achieve a more comprehensive video understanding across the entire spatiotemporal dimension.

\bibliographystyle{aaai22}
\bibliography{neurips_2023}

\end{document}